\documentclass{article}

\PassOptionsToPackage{numbers}{natbib}

\usepackage[main, final]{neurips_2026}

\usepackage[utf8]{inputenc} 
\usepackage[T1]{fontenc}    
\usepackage{hyperref}       
\usepackage{url}            
\usepackage{booktabs}       
\usepackage{amsfonts}       
\usepackage{nicefrac}       
\usepackage{microtype}      
\usepackage{xcolor}         
\usepackage{tcolorbox}      
\usepackage{amsmath}
\usepackage{wrapfig}
\tcbuselibrary{skins, breakable}    
\usepackage[normalem]{ulem} 
\usepackage{soul}           
\usepackage{multirow}
\usepackage{algorithm}
\usepackage{algorithmic}
\usepackage{caption} 
\definecolor{icebluebox}{RGB}{237, 242, 249} 
\definecolor{icebluetext}{RGB}{41, 105, 176} 
\definecolor{hlred}{RGB}{255, 214, 214}      
\newcommand{\gdel}[1]{\textcolor{red}{\sout{#1}}}
\newcommand{\kkd}[2]{{\setlength{\fboxsep}{1.5pt}\colorbox{hlred}{\textbf{#1}\textbf{\textcolor{red}{\sout{#2}}}}}}
\newcommand{\kdel}[1]{{\setlength{\fboxsep}{1.5pt}\colorbox{hlred}{\textbf{\textcolor{red}{\sout{#1}}}}}}

\newtcolorbox{prettybox}{   
    enhanced,
    breakable,
    colback=white, 
    colframe=gray!40,           
    boxrule=0.6pt,              
    arc=2.5mm,                  
    left=2mm, right=2mm, top=2mm, bottom=2mm, 
    boxsep=0pt
}

\newtcolorbox{contextbox}{      
    enhanced,
    colback=icebluebox,        
    colframe=icebluebox!80!gray,
    boxrule=0.5pt,
    arc=1.5mm,                  
    left=2mm, right=2mm, top=2mm, bottom=2mm,
    boxsep=0pt
}

\title{\ours{}: Weight-Space Compensation of KV Cache}

\author{Chanryeol Lee\hspace{1em}Chanhyuk Lee\hspace{1em}Yeonwoo Choi\hspace{1em}Donggyun Kim\textsuperscript{\dag}\hspace{1em} Seunghoon Hong\textsuperscript{\dag} \\
KAIST \\
\texttt{\{lcy9442, chan3684, lotus\_68, kdgyun425, seunghoon.hong\}@kaist.ac.kr}
}

\usepackage{enumitem}
\usepackage{graphicx}

\newcommand{\ours}{PatchKV}

\begin{document}

\maketitle

\renewcommand{\thefootnote}{\dag}
\footnotetext{Equal advising}
\renewcommand{\thefootnote}{\arabic{footnote}}
\setcounter{footnote}{0}

\begin{abstract}
    Long-context inference with Large Language Models (LLMs) is bottlenecked by the linearly growing memory of the key-value (KV) cache.
    Existing compression methods reduce the cache through token eviction or approximation, but degrade sharply at aggressive compression budgets.
    We propose \ours{}, a training-free framework that compensates KV cache compression methods by carrying part of the context in the model's weights.
    \ours{} pairs an off-the-shelf compressed KV cache with a context-specific weight patch, which is computed once at context-loading time and served for downstream queries for the context. 
    The weight patch is derived in closed form via ridge regression, by aligning the block-wise activations of context-derived reference query tokens under the full cache and the compressed cache.
    Once merged into the model, the patch leaves the forward graph and per-query inference cost unchanged in the single-context, multi-query setting.
    Across long-context QA (SCBench with up to 170K tokens, SQuAD, NIAH) and math (GSM8K) benchmarks on three model architectures, \ours{} consistently improves cache compression methods, suggesting an alternative direction to compensate them at aggressive budgets. Code is available at \url{https://github.com/cusasak/PatchKV}.
\end{abstract}

\section{Introduction}
\label{sec:intro}

Many practical applications of Large Language Models (LLMs) follow a single-context, multi-query pattern~\citep{li2025scbench}, where a long context is read once and then queried repeatedly.
This includes scenarios such as a code assistant grounding many edits in the same repository, a document analyst asking multiple questions about the same report, and a customer-support agent handling each conversation against a fixed knowledge base.
In these settings, the cost of processing the context can be amortized across subsequent queries, while the latency and memory cost of each query directly affect serving efficiency.

To avoid recomputing context-token activations for every query, modern LLMs store their key and value states as a KV cache during generation, whose size grows linearly with context length~\citep{pope2023efficiently, zhang2023h2o}.
At long context lengths, this cache can dominate the memory footprint and memory-bandwidth cost of serving additional queries, making KV cache reduction a central tool for efficient long-context inference~\citep{liu2024kivi, hooper2024kvquant}.
A straightforward way to reduce this bottleneck is local context windowing, which retains only the most recent context tokens and discards distant ones.
However, simply discarding distant context tokens can degrade downstream performance, since subsequent queries may require information contained in the discarded portion of the context~\citep{xiao2024efficient, xiao2024infllm}.

To reduce KV cache memory without simply truncating the context to a local window, recent work compresses the KV cache more selectively.
Token-eviction methods discard cache entries judged less important by attention scores or other heuristics~\citep{zhang2023h2o, xiao2024efficient, li2024snapkv, kim2025kvzip}, and cache-approximation methods construct a smaller surrogate cache that approximates the full one~\citep{eyuboglu2025cartridges, sun2025shadowkv, zweiger2026fast}.
These methods differ in how they select or synthesize cache entries, but they primarily rely on the compressed cache to store context information after compression.
This design is effective across many settings, but its accuracy can degrade as the cache budget becomes more aggressive~\citep{tang2025razorattention, sharma2025minikv}.
In particular, a small fixed cache may fail to retain some facts, long-range dependencies, or query-relevant details needed by future queries, especially when the relevant content is difficult to anticipate at compression time.
This motivates a complementary mechanism that can recover missing context information through a channel other than the compressed cache itself.

Meanwhile, recent theoretical studies suggest that the model weights can be used as such a secondary channel for storing context information~\citep{dherin2025learning, goldwaser2026equivalence, mazzawi2025transmuting}.
The results show that, under simplified settings, the behavior of attending to a fixed context can be replaced by context-dependent weight updates of the model.
This perspective suggests that context information can be expressed through two coupled mechanisms: the KV cache and the model weights.
However, existing work has largely focused on theoretical structure or toy settings, often with fixed query tokens and the retained context restricted to a subset of the original tokens.
This leads us to explore more practical usage of the context-to-weight conversion for KV cache compression: \textbf{can a weight correction purely driven by the context recover part of the information lost by compression?}

We answer this question with \ours{}, a weight-based approach for compensating KV cache compression in a single-context, multi-query setting.
Given a context, \ours{} first obtains an off-the-shelf compressed KV cache.
It then computes a context-specific \emph{weight patch} by comparing two intermediate activations on context-derived reference sequences: a full-cache teacher and a compressed-cache student.
For each transformer block, we solve a closed-form ridge regression that aligns the student's block activations with those of the teacher, and we apply the resulting weight patch to the MLP output projection.
Once the patch is merged into the weights for the active context, the generation-time forward graph is unchanged, thus all additional computation is paid once at context-loading time and amortized across later queries.

We empirically evaluate \ours{} with three LLM backbones and two KV cache compression methods on a benchmark suite spanning long-context QA (SCBench~\citep{li2025scbench}, SQuAD~\citep{rajpurkar-etal-2016-squad}, NIAH~\citep{needle}) and math reasoning (GSM8K~\citep{cobbe2021training}).
Across token-eviction and cache-approximation compression methods, \ours{} consistently improves performance at aggressive cache budgets (\emph{e.g.,} $<20\%$) and recovers a substantial fraction of the accuracy gap between compressed-cache and full-cache inference.
The results show that \ours{} can partially restore missing information in context through the weight patch, suggesting a complementary direction to memory-efficient generation in long-context settings.

Our contributions are as follows.
\vspace{-0.2cm}
\begin{itemize}[leftmargin=0.5cm]
    \item \textbf{Weight-space compensation of KV cache compression.}
    We formulate context-specific weight patching as a complementary mechanism to KV cache compression, allowing context information lost by aggressive compression to be partially recovered through the model weights.

    \item \textbf{A closed-form patch construction.}
    We derive a gradient-free block-wise ridge regression that patches MLP output projections by matching compressed-cache student activations to full-cache teacher activations, without access to downstream queries.

    \item \textbf{Compressor-agnostic validation.}
    We instantiate the framework on top of both token-eviction and cache-approximation methods and evaluate its compensating behavior through extensive empirical studies in long-context QA and math benchmarks.
\end{itemize}
\section{Method}
\label{sec:method}

\subsection{Problem Formulation}
\label{sec:problem_formulation}

We consider tasks where a single long context is prefilled once and reused across many downstream queries, such as multi-turn document QA, repeated retrieval over a database-like context, or iterative analysis of a long document.
In this regime, the one-time work at context loading can be amortized over many future queries.
In contrast, per-query cost such as latency, memory footprint, and memory bandwidth remain directly exposed at serving time.
Since each query must attend to the long context to produce context-dependent responses, preprocessing the context so that per-query access is efficient becomes a major practical concern in the single-context, multi-query setting.

A common strategy for efficient inference is to manage a KV cache~\citep{haoyang2025survey} of the prefilled context.
Given a context token sequence $c = (c_1, \ldots, c_n)$ of length $n$, a transformer-based LLM first runs a prefill pass to produce a full KV cache $\mathcal{C} = \{(K_\ell, V_\ell)\}_{\ell=1}^{L}$, where $K_\ell$ and $V_\ell$ are the key and value activations of all context tokens at attention layer $\ell$.
The cache is retained in running memory so that subsequent queries can attend to the context during autoregressive generation without recomputing the prefill.
However, as the memory footprint of $\mathcal{C}$ scales linearly with $n$, long-context serving often requires reducing the cache before answering future queries to avoid memory overflow.

\begin{figure}
    \centering\centerline{\includegraphics[width=1.00\textwidth]{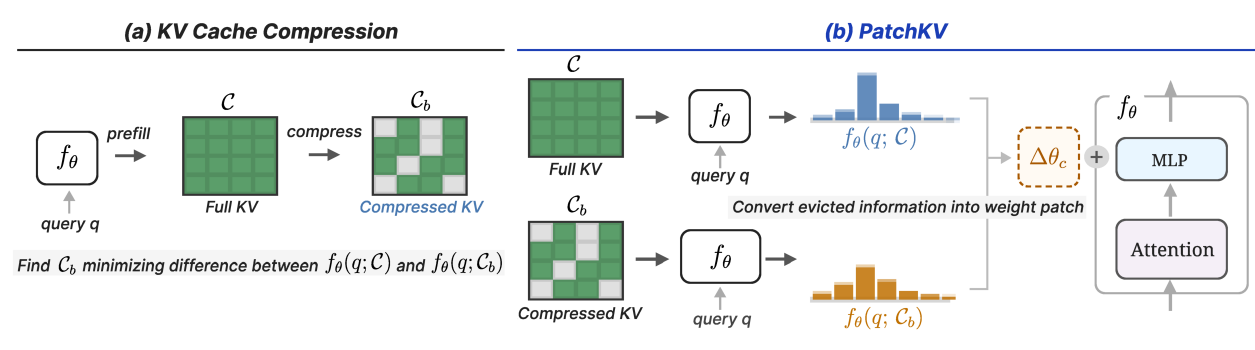}}
    \vspace{-0.1cm}
    \caption{
    Overview of \ours{}.
    (a) Standard KV cache compression stores context information only in a compressed cache.
    (b) \ours{} pairs the compressed cache with a context-specific patch to the MLP output projections, compensating for part of the full-cache behavior lost during compression.
    }
    \label{fig:overview}
\end{figure}

\textbf{Background: KV Cache Compression.}
Recent KV cache compression approaches reduce the cache memory at context-prefill time either by evicting cache entries judged less important~\citep{li2024snapkv, kim2025kvzip} or by constructing a smaller cache that approximates the full one~\citep{zweiger2026fast} (see Fig.~\ref{fig:overview}(a)).
Let $f_\theta(q;\mathcal{C})$ denote the output distribution of a model with parameters $\theta$ when a query sequence $q=(q_1,\ldots,q_m)$ attends to cache $\mathcal{C}$, and let $\mathcal{L}(\cdot,\cdot)$ be a discrepancy between output distributions.
An idealized cache-compression objective can be written as
\begin{equation}
    \min_{\mathcal{C}_b:\,|\mathcal{C}_b| \leq b}
    \;
    \underset{q \sim p_\text{down}(q|c)}{\mathbb{E}}
    \Big[
        \mathcal{L}\big(
        f_\theta(q;\mathcal{C}),
        f_\theta(q;\mathcal{C}_b)
        \big)
    \Big],
    \label{eq:compression}
\end{equation}
where $\mathcal{C}_b$ is the compressed cache, $b$ is the cache budget, and $p_\text{down}(q|c)$ is the downstream query distribution conditioned on context $c$.
Since $p_\text{down}(q|c)$ is unknown at compression time, a common practice is to approximate it using proxy queries derived from the context~\citep{li2024snapkv, kim2025kvzip, eyuboglu2025cartridges, zweiger2026fast}.
Once constructed, $\mathcal{C}_b$ is reused across all downstream queries, reducing the per-query serving cost by a factor of $b/n$.

Cache compression methods preserve context information solely through $\mathcal{C}_b$, where $b$ must be carefully chosen to balance accuracy against efficiency.
At aggressive budgets $b$, this cache-only bottleneck can drop details needed by future queries, and recovering them by increasing $b$ comes at a direct per-query cost.
We therefore propose to augment the compressed cache with a second channel: a patch to the model weights.

\textbf{Our Approach: Compensating Compression Error via Weight Patching.}
To complement the compressed cache $\mathcal{C}_b$ without increasing the budget $b$, we introduce a context-specific weight patch $\Delta\theta_c$ that is constructed in the context-prefill stage and then merged into the base weights $\theta$ to serve downstream queries.
Ideally, the patch is designed to narrow the remaining gap between full-cache inference and compressed-cache inference on the downstream queries:
\begin{equation}
    \min_{\Delta\theta_c}
    \;
    \underset{q \sim p_\text{down}(q|c)}{\mathbb{E}}
    \Big[
        \mathcal{L}\big(
        f_\theta(q;\mathcal{C}),
        f_{\theta+\Delta\theta_c}(q;\mathcal{C}_b)
        \big)
    \Big].
    \label{eq:ideal_objective}
\end{equation}
As in KV cache compression, one practical challenge is that $p_\text{down}(q|c)$ is unavailable at the patch construction time.
Therefore, we follow their practices to approximate it with a finite reference set $\mathcal{Q}_c$ derived from the context $c$:
\begin{equation}
    \min_{\Delta\theta_c}
    \;
    \underset{q \sim \mathcal{Q}_c}{\mathbb{E}}
    \Big[
        \mathcal{L}\big(
        f_\theta(q;\mathcal{C}),
        f_{\theta+\Delta\theta_c}(q;\mathcal{C}_b)
        \big)
    \Big].
    \label{eq:objective}
\end{equation}
Sec.~\ref{sec:reference} describes specific reference queries used in our experiments, while Sec.~\ref{sec:closed-form} and~\ref{sec:efficient_computation} describe a specific instantiation of this objective and practical techniques for long-context problems.
The overall algorithm is provided in App.~\ref{appendix:algorithm}.

The key benefit of the weight-space compensation for KV cache compression is that it shifts the cost of recovery from per-query to a one-time prefill stage.
Once $\Delta\theta_c$ is merged into the base weights $\theta$, the patched model uses the same forward graph as the base model and introduces no additional per-token computation during generation.
The patched model can then fully inherit the memory and latency improvements of the compressed KV cache at more aggressive budgets $b$, without paying the per-query cost that a larger $b$ would otherwise incur.
Moreover, our formulation is agnostic to the choice of $\mathcal{C}_b$.
Any off-the-shelf KV cache compression method that operates at the prefill stage can be plugged in, making the weight-space patch complementary to existing token-space compression.

\subsection{Closed-Form Weight Patch Derivation}
\label{sec:closed-form}

To obtain the weight patch minimizing the approximated objective Eq.~\eqref{eq:objective} without performing heavy training on the model parameters, we borrow ideas from recent studies in context-to-weight conversion~\citep{dherin2025learning, goldwaser2026equivalence, mazzawi2025transmuting}.
Prior work suggests a mechanism to convert the effect of context in transformer architectures into weight updates on the post-attention MLP layers.
We adapt this idea to our long-context, multi-query setting as a complement to KV cache compression.

\textbf{MLP activation matching.}
Motivated by the prior work~\citep{dherin2025learning, goldwaser2026equivalence, mazzawi2025transmuting}, we instantiate Eq.~\eqref{eq:objective} as a block-wise activation matching objective on reference queries $q \in \mathcal{Q}_c$, constructing a separate patch for each MLP block.
For clarity, we present the derivation for a single block and drop the block index throughout.
At each block, we write the block output as the sum of the MLP output $W h$ and a residual term $z$.
Here $W \in \mathbb{R}^{d \times d_{\text{ff}}}$ is the MLP output projection (the down-projection, in SwiGLU variants), $h \in \mathbb{R}^{d_{\text{ff}}}$ is the post-activation intermediate, and $z \in \mathbb{R}^{d}$ is the attention output, where $d$ and $d_{\text{ff}}$ denote the hidden and MLP intermediate dimensions.
KV cache compression perturbs the hidden states entering each block and therefore changes both terms.
For each reference token $i$ in some $q \in \mathcal{Q}_c$, we obtain teacher activations $(h_i, z_i)$ and student activations $(h'_i, z'_i)$ from teacher-forced forward passes over $q$ with caches $\mathcal{C}$ and $\mathcal{C}_b$ respectively.

We seek $\Delta W$ such that the patched compressed-cache block output approximates the corresponding full-cache block output for reference tokens:
\begin{equation}
    W h_i + z_i = (W + \Delta W) h'_i + z'_i.
    \label{eq:per-token-match}
\end{equation}
Equivalently, the patch uses the compressed-cache MLP features $h'_i$ as a basis for correcting the full-cache versus compressed-cache block-output discrepancy.
We derive patches sequentially across blocks: when solving for $\Delta W$ at block $\ell$, the upstream blocks $1,\ldots,\ell-1$ are already patched, so $h'_i$ and $z'_i$ reflect activations from a model that has itself been partially corrected.
This greedy formulation is consistent with how compression errors propagate at inference time, where each block receives the residual stream produced by the corrected blocks beneath it.

\textbf{Closed-form solution with ridge regression.}
Rearranging Eq.~\eqref{eq:per-token-match} gives the per-token condition: \begin{equation}
    \Delta W h'_i = t_i, \quad\text{where}\quad t_i = W(h_i - h'_i) + (z_i - z'_i).
    \label{eq:per_token_condition}
\end{equation}
For a single token, this condition is underdetermined and admits infinitely many token-specific solutions~\citep{dherin2025learning, goldwaser2026equivalence}, none of which necessarily generalizes beyond that token.
To obtain a single context-specific but query-agnostic patch shared across all reference tokens, we stack $\{h'_i\}$ and $\{t_i\}$ as rows of $H' \in \mathbb{R}^{N_{\mathrm{ref}} \times d_{\text{ff}}}$ and $T \in \mathbb{R}^{N_{\mathrm{ref}} \times d}$, where $N_{\mathrm{ref}}$ is the total number of tokens across all sequences in $\mathcal{Q}_c$.
We then solve the ridge-regularized least-squares problem as in \citet{mazzawi2025transmuting}:
\begin{equation}
    \min_{\Delta W}\; \big\| \Delta W H^{\prime\top} - T^{\top} \big\|_F^2 + \lambda \|\Delta W\|_F^2,
    \label{eq:ridge}
\end{equation}
where the ridge coefficient $\lambda > 0$ ensures the system is well-conditioned and discourages overly large patches that overfit the reference query set.
Because the scale of $H'^\top H'$ varies substantially across layers, we introduce an adaptive ridge scaling as
\begin{equation}
    \lambda = \lambda_0 \cdot \frac{\|H'^\top H'\|_F^2}{\mathrm{tr}(H'^\top H')},
    \label{eq:ridge_scaling}
\end{equation}
where $\lambda_0$ is a single hyperparameter shared across layers.
This normalization makes $\lambda$ comparable across layers and architectures, requiring no per-layer tuning.
The ridge objective (Eq.~\eqref{eq:ridge}) admits the closed-form solution
\begin{equation}
    \Delta W = T^{\top}H' \big( H^{\prime\top}H' + \lambda I \big)^{-1}.
    \label{eq:ridge-solution}
\end{equation}
After solving $\Delta W$ for each block, we merge them into the base model to serve downstream queries.

\subsection{Efficient Computation for Long Reference Sequences}
\label{sec:efficient_computation}

Naively applying Eq.~\eqref{eq:ridge-solution} requires materializing all per-token activations $\{(h'_i, t_i)\}$ across the full reference query in a single forward pass and stacking them into matrices $H'$ and $T$, incurring memory cost that grows linearly with the total number of query tokens $N_\mathrm{ref}$.
This becomes prohibitive in our long-context setting, where reference queries can be as long as the context itself (\emph{e.g.,} repeat-context introduced in Sec.~\ref{sec:reference}).
We address this in two stages.

\textbf{Chunked query encoding.}
First, we process the reference sequence in fixed-size chunks at the forward-pass level.
Each chunk is run through the model independently, so peak activation memory is determined by the chunk size rather than the full sequence length.
This introduces an approximation, since cross-chunk attention is dropped.
However, our reference queries are designed so that the relevant dependencies are local, \emph{e.g.}, reconstructing specific local segments of the context (Sec.~\ref{sec:reference}), which keeps this approximation mild.
We evaluate its effect in Sec.~\ref{sec:ablation}.

\textbf{Efficient regression with sufficient statistics.}
Second, even with chunked forward passes, naively stacking the resulting activations $\{(h'_i, t_i)\}$ across all chunks would still incur memory linear in the total number of tokens $N_\mathrm{ref}$.
The key observation is that, at any given block, the closed-form solution depends on reference tokens only through two sufficient statistics,
\begin{equation}
    S_H := H^{\prime\top} H' \in \mathbb{R}^{d_{\text{ff}} \times d_{\text{ff}}},
    \qquad
    S_T := T^{\top} H' \in \mathbb{R}^{d \times d_{\text{ff}}},
    \label{eq:stats}
\end{equation}
both of which are sums of per-token outer products.
We therefore accumulate $(S_H, S_T)$ as running sums starting from zero matrices.
For each chunk $j$, we compute its outer-product contributions $S_H^{(j)} = H_j'^\top H_j'$ and $S_T^{(j)} = T_j^\top H_j'$ where $H_j'$ and $T_j$ denote the stacked activations of $\{h_i'\}$ and $\{t_i\}$ within chunk $j$, add them to the running totals, and discard the activations of chunk $j$ before moving on.
This makes peak activation memory independent of $N_\mathrm{ref}$ and depends only on the chunk size $L_\mathrm{chunk}$ and on the size of the running statistics.

Once all chunks have been processed, we solve a single ridge regression $\Delta W = S_T(S_H + \lambda I)^{-1}$
per block, identical in form to Eq.~\eqref{eq:ridge-solution} but using statistics accumulated incrementally rather than materialized all at once.

\subsection{Reference Query Construction}
\label{sec:reference}

As discussed in Sec.~\ref{sec:problem_formulation}, we do not have access to ground-truth downstream query distribution $p_\text{down}(q|c)$ and thus need an approximation.
Following prior KV cache compression work, we approximate $p_\text{down}(q|c)$ with a finite reference set $\mathcal{Q}_c = \{q^{(1)},\ldots,q^{(r)}\}$.
The reference set is derived purely from the context itself without any downstream labels. We consider three reference-construction strategies.
\vspace{-0.2cm}
\begin{itemize}[leftmargin=0.5cm]
    \item \textbf{Repeat-context.}
    Following \citet{kim2025kvzip}, we construct teacher-forced reconstruction sequences from the context itself.
    For short contexts, we use a repeat prompt $P$ such as \texttt{Repeat the previous context.} followed by the context tokens, forming $\mathcal{Q}_c=\{[P;c]\}$.
    For long contexts, we split $c$ into $N$ chunks $\{c^{(j)}\}_{j=1}^{N}$ with chunk-specific prompts $P^{(j)}$, \emph{e.g.,} \texttt{Repeat the previous context after <chunk anchor>}, giving $\mathcal{Q}_c = \{[P^{(j)};c^{(j)}]\}_{j=1}^{N}$.
    All tokens are teacher-forced rather than generated.
    This strategy has no sampling overhead and encourages reference activations to cover fine-grained facts across the context.

    \item \textbf{Self-study.}
    Following \citet{eyuboglu2025cartridges}, we prompt the full-cache base model with a small set of fixed, context-agnostic instructions
    $I^{(1)},\ldots,I^{(M)}$, \emph{e.g.,} \texttt{Aggregate all key facts mentioned in the context}.
    The model generates an answer $a^{(j)}$ for each instruction using the full cache, and we use the resulting pairs as teacher-forced reference sequences $\mathcal{Q}_c = \{[I^{(1)};a^{(1)}],\ldots,[I^{(M)};a^{(M)}]\}$.
    Since the answers $a^{(j)}$ are generally short, each sequence $[I^{(j)};a^{(j)}]$ is treated as a single chunk.
    Compared with repeat-context, self-study may better cover higher-level summaries and reasoning patterns, but it adds generation cost at context loading and risks larger discrepancy from downstream queries.

    \item \textbf{Joint.}
    We pool the reference queries from repeat-context and self-study into a single set $\mathcal{Q}_c$.
    This provides a conservative approximation to $p_\text{down}(q|c)$: self-study aggressively explores plausible downstream queries, while repeat-context completely covers the context.
\end{itemize}

\section{Experiments}
\label{sec:experiments}

Our experiments focus on evaluating whether \ours{} can compensate for the information lost during KV compression.
We first evaluate the effect of \ours{} over KV cache compression methods, and then analyze the behavior of the weight patch through ablations and controlled experiments.

\begin{figure}
    \centering
    \includegraphics[width=0.96\linewidth]{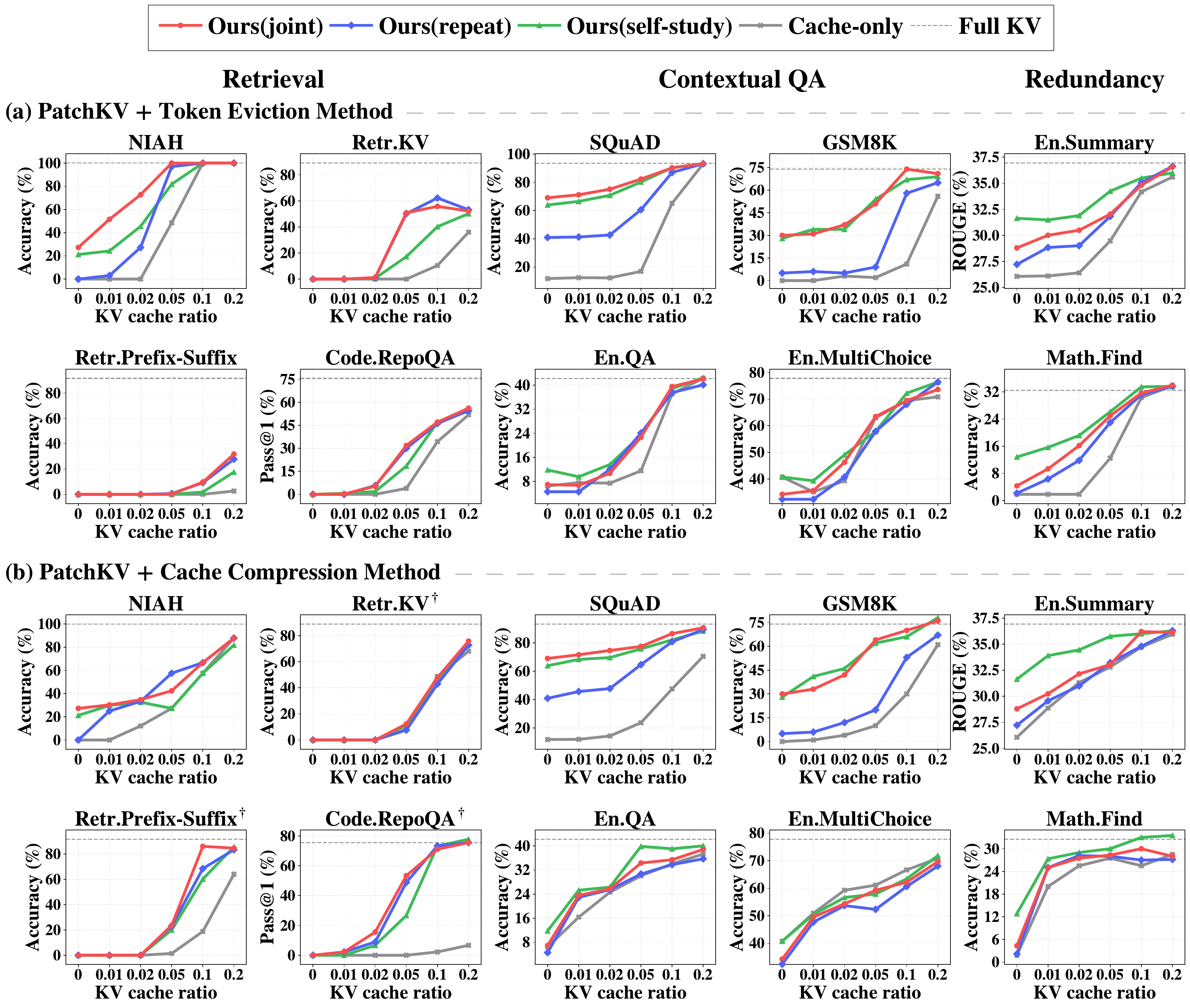}
    \caption{Benchmark results using Qwen2.5-7B-1M with \ours{} applied on top of \textbf{(a)} an eviction based method (KVzip) and \textbf{(b)} an approximation based method (Attention Matching). In \textbf{(b)}, $^{\dagger}$ marks shortened (tiny) task variants for SCBench retrieval tasks.
    }   
    \label{fig:main-kvzip}
    \vspace{-0.2cm}
\end{figure}

\subsection{Setup}
\label{sec:setup}

\textbf{KV cache compression methods.}
We consider two off-the-shelf KV cache compression methods to which \ours{} is applied: KVzip~\cite{kim2025kvzip} and Attention Matching~\cite{zweiger2026fast}, representing token-eviction and cache-approximation methods, respectively.
KVzip first prefills the context and scores each context KV pair via a teacher-forced reconstruction pass with the repeat-context prompt described in Sec.~\ref{sec:reference}.
KV pairs that receive low maximum cross-attention during reconstruction are then evicted according to the target cache budget.
Attention Matching, in contrast, constructs a compacted cache by solving a regression problem that matches full-cache attention outputs using compacted value activations and additional bias terms while evicting the key activations, using both the repeat-context and self-study prompts described in Sec.~\ref{sec:reference}.
In both cases, \ours{} is applied after compression, and the resulting patched model is then used to serve downstream queries.

\textbf{Benchmarks and models.}
We evaluate on benchmarks spanning three modes of long-context utilization: exact retrieval, contextual reasoning, and robustness to redundant context.
Specifically, we use SQuAD~\cite{rajpurkar-etal-2016-squad}, GSM8K~\cite{cobbe2021training}, needle-in-a-haystack (NIAH)~\cite{needle}, and seven benchmarks from SCBench~\cite{li2025scbench}, which together span context lengths up to 170K tokens.
SCBench additionally includes \texttt{Retrieve.Multihop} and \texttt{ICL.ManyShot}, which we report separately because they exhibit qualitatively different behavior under compression; see App.~\ref{app:dataset_selection}.
We follow the official metric of each benchmark: exact-match accuracy for retrieval and QA tasks, answer accuracy for GSM8K, Pass@1 for code tasks, and ROUGE~\cite{lin2004rouge} for summarization.

Results are reported across KV cache budget ratios $b/n$ in the aggressive regime, \emph{i.e.} at most $20\%$.
For the Attention Matching results in Fig.~\ref{fig:main-kvzip}(b), we evaluate shortened variants of the SCBench retrieval tasks, as both Attention Matching and Attention Matching with \ours{} achieve zero accuracy on retrieval tasks under the evaluated cache budgets. We report the original task results in App.~\ref{app:dataset_selection}.
To validate \ours{} across different model architectures, we use three base LLMs: Qwen2.5-7B-1M~\cite{qwen2025qwen25technicalreport}, Qwen3-4B~\cite{yang2025qwen3}, and LLaMA3.1-8B~\cite{grattafiori2024llama}.
Implementation details are provided in App.~\ref{appendix_setup}.

\subsection{Main Results}\label{sec:main_results}

Fig.~\ref{fig:main-kvzip} reports the effect of \ours{} applied on top of KVzip and Attention Matching, with three variants corresponding to the reference query strategies introduced in Sec.~\ref{sec:reference}.
The most pronounced gains occur in the aggressive cache regime ($b/n$ between 0.02 and 0.1), where the cache-only baselines drop sharply while \ours{} substantially narrows the gap to full-cache inference. 
In this regime, \ours{} often matches or exceeds the cache-only baseline obtained with a noticeably larger budget.
This suggests that part of the missing context information can be recovered through the weight patch rather than through additional cache capacity.

The effect is most striking at the extreme end of the budget range:
on information-dense tasks such as SQuAD and GSM8K, \ours{} retains nontrivial accuracy even when no KV cache remains.
On SQuAD, in particular, \ours{} recovers up to $73\%$ of full-KV accuracy with no cache, whereas the cache-only baseline collapses to near zero.
Fig.~\ref{fig:quali-1} illustrates this behavior on a representative SQuAD example: KVzip evicts most of the context surrounding the answer token ``Leprechaun,'' causing the cache-only model to produce the incorrect answer ``Fighting Irish,'' whereas applying \ours{} on top of the same compressed cache recovers the correct answer ``Notre Dame Leprechaun.''
Additional qualitative examples are provided in App.~\ref{appendix:quan_exp}.

These improvements hold across both KVzip (token eviction) and Attention Matching (cache approximation), suggesting that the weight-patching mechanism is largely orthogonal to the choice of compression strategy.
Interestingly, the three reference query strategies exhibit clearly different profiles: self-study is most effective on contextual QA and redundancy-heavy tasks, while repeat-context and the joint variant are stronger on retrieval-heavy tasks.
This pattern is consistent with the design of each strategy where self-study emphasizes higher-level summaries and reasoning patterns, whereas repeat-context provides fine-grained coverage of every context segment. Overall, \ours{} delivers consistent gains across context lengths, compression methods, and task types, with the largest benefits realized in the regime where compression methods strongly degrade.

\begin{table}[t]
  \centering
  \caption{Perplexity ($\downarrow$) on SQuAD across compression ratios. ref. stands for reference query.}
  \vspace{0.2cm}
  \label{tab:kv-perplexity}
  \footnotesize
    \renewcommand{\arraystretch}{1.2}
    \renewcommand{\aboverulesep}{0pt}
    \renewcommand{\belowrulesep}{0pt}
    \setlength\tabcolsep{6pt}
  \begin{tabular}{l|cc|cc|cc|cc|cc}
    \toprule
    \textbf{Ratio} & \multicolumn{2}{c|}{\textbf{Full KV}} & \multicolumn{2}{c|}{\textbf{None}} & \multicolumn{2}{c|}{\textbf{repeat-context}} & \multicolumn{2}{c|}{\textbf{self-study}} & \multicolumn{2}{c}{\textbf{joint}} \\
    \cmidrule(lr){2-3} \cmidrule(lr){4-5} \cmidrule(lr){6-7} \cmidrule(lr){8-9} \cmidrule(lr){10-11}
    & ref. & test & ref. & test & ref. & test & ref. & test & ref. & test \\
    \midrule
    0.20 & 1.00 & 8.79 & 1.00 & 18.16 & 1.00 & \textbf{12.24} & 1.00 & 17.01 & 1.00 & 17.07 \\
    0.10 & 1.00 & 8.79 & 1.10 & 17.16 & 1.00 & \textbf{14.03} & 1.03 & 16.51 & 1.00 & 16.60 \\
    0.05 & 1.00 & 8.79 & 1.81 & 123.36 & 1.00 & 21.08 & 1.10 & 18.90 & 1.00 & \textbf{18.84} \\
    0.02 & 1.00 & 8.79 & 6.18 & 180.63 & 1.00 & 32.82 & 1.39 & \textbf{28.34} & 1.00 & 30.74 \\
    \bottomrule
  \end{tabular}
  \vspace{-0.2cm}
\end{table}

\begin{figure}
    \centering
    
    \begin{prettybox}\scriptsize
        \textbf{Question:} What type of mascot do the Notre Dame sport teams have?
        \hfill \textcolor{gray}{\scriptsize \textsf{[SQuAD]}}
    
        \begin{contextbox}
        \textbf{Pruned Context:} [...] The official colors of Notre\gdel{ Dame} are Navy Blue and Gold Rush which are worn in competition by its athletic teams.\gdel{ In} addition,\gdel{ the} color green is often worn because of the Fighting Irish nickname.\gdel{ The} Notre Dame \kkd{Lepre}{chaun} is the mascot of the athletic teams. Created by Theodore W. Drake in\gdel{ 19}64, the \kkd{le}{prechaun} was first used on\gdel{ the} football pocket schedule and later on\gdel{ the} football program covers. The \kkd{le}{prechaun} was featured on the cover of Time in November \gdel{196}4 and gained national exposure.
        \end{contextbox}
    
        {\color{black!20}\hrule height 0.8pt} 
        \vspace{1.5mm} 
    
        \noindent\textbf{Full KV:} The Notre Dame sport teams have the Notre Dame Leprechaun as their mascot. \hfill \textcolor{gray}{\scriptsize \textsc{Correct}}
    
        \vspace{1.5mm}
        {\color{black!10}\hrule height 0.5pt} 
        \vspace{1.5mm}
    
        \noindent\textbf{Pruned KV (w/o patch):} The Notre Dame sport teams have a \textcolor{red}{Fighting Irish} mascot. \hfill \textcolor{red}{\scriptsize \textsc{Incorrect}}
    
        \vspace{1.5mm}
        {\color{black!10}\hrule height 0.5pt} 
        \vspace{1.5mm}
    
        \noindent\textbf{\ours{} (Ours):} The Notre Dame sport teams have a mascot called the Notre Dame \textcolor{icebluetext}{Leprechaun}. \hfill \textcolor{icebluetext}{\scriptsize \textsc{Correct}}
    \end{prettybox}
    \vspace{-0.2cm}
    \caption{Qualitative example of \ours{} over KVzip on SQuAD. Evicted tokens shown in red.}
    \label{fig:quali-1}
    \vspace{-0.2cm}
\end{figure}

\subsection{Analysis}

\textbf{Generalization ability of reference queries.}
We examine how well a weight patch constructed from the reference query set $\mathcal{Q}_c$ generalizes to downstream queries $q \sim p_\text{down}(q|c)$.
On SQuAD, we measure teacher-forced perplexity under both patched and non-patched models on two sets of tokens: (i) the reference queries and (ii) the gold answer tokens conditioned on each downstream query.
The results are reported in Tab.~\ref{tab:kv-perplexity}.
The patched model nearly perfectly fits the reference queries, matching the perplexity obtained with the full KV cache.
This indicates that the block-wise activation matching used to derive the ridge solution (Eq.~\eqref{eq:ridge-solution}) is an effective proxy for the distribution-level reference objective (Eq.~\eqref{eq:objective}) introduced in Sec.~\ref{sec:problem_formulation}.
On downstream test queries, the patched model does not fully close the gap to full-KV perplexity, particularly in the low-cache regime.
Nevertheless, it consistently narrows the perplexity gap between the full and compressed KV caches across all budgets, indicating that the patch generalizes beyond the reference set rather than merely memorizing it. We observe that patches built from self-study based reference queries generalize better than repeat-context in majority.

\textbf{Weight patch complements the compressed cache.}
Next, we test whether the patch captures information lost by compression rather than information redundant with the compressed KV cache.
On SQuAD, we first evict 80\% of KV pairs via KVzip and randomly split the remaining 20\% into two disjoint subsets, $\mathcal{C}_A$ and $\mathcal{C}_B$.
We then construct two weight patches $\Delta W_A$ and $\Delta W_B$ using only $\mathcal{C}_A$ and $\mathcal{C}_B$, respectively, and measure performance under three configurations: (i) the compressed cache alone ($\mathcal{C}_A$ or $\mathcal{C}_B$), (ii) each compressed cache with its corresponding weight patch ($\mathcal{C}_A+\Delta W_A$, $\mathcal{C}_B+\Delta W_B$), and (iii) each compressed cache with the patch derived from the disjoint subset ($\mathcal{C}_A+\Delta W_B$, $\mathcal{C}_B+\Delta W_A$).

\begin{wraptable}{r}{0.3\textwidth} 
  \small
  \vspace{-1.2em}
  \centering
  \caption{\textbf{Cross-evaluation of patches built from disjoint cache subsets.} Rows: weight patch applied at inference. Columns: compressed cache used at inference.}
  \vspace{-0.1cm}
  \label{tab:kv-patch-orthogonal}
  \begin{tabular}{lcc}
    \toprule
    & \multicolumn{2}{c}{\textbf{Eval cache}} \\
    \cmidrule(lr){2-3}
    \textbf{Patch} & $\mathcal{C}_A$ & $\mathcal{C}_B$ \\
    \midrule
    None            & 54.95 & 57.74 \\
    $\Delta W_A$    & \textbf{85.39} & 66.60 \\
    $\Delta W_B$    & 65.63 & \textbf{88.09} \\
    \bottomrule
  \end{tabular}
\end{wraptable}

The results are reported in Tab.~\ref{tab:kv-patch-orthogonal}.
Since $\mathcal{C}_A$ and $\mathcal{C}_B$ are randomly drawn from the same context, the two caches have similar capacity, as reflected by their comparable no-patch performance (top row).
Each patch performs best when paired with its source cache (diagonal entries) and degrades when paired with the disjoint subset's patch (off-diagonal entries), while still outperforming the cache-only baseline.
This intermediate behavior is consistent with the design of the weight-patching objective (Eq.~\eqref{eq:objective}): both patches are trained to fill the gap between full-cache and compressed-cache behavior, so they share the bulk of that gap (the evicted $80\%$ of the context) but differ in the smaller portion that corresponds to the other subset.
The shared component transfers under swapping and explains the gain over the cache-only baseline, whereas the subset-specific component is recovered only when the patch is paired with its source cache.

\begin{figure}
    \centering
    \includegraphics[width=\linewidth]{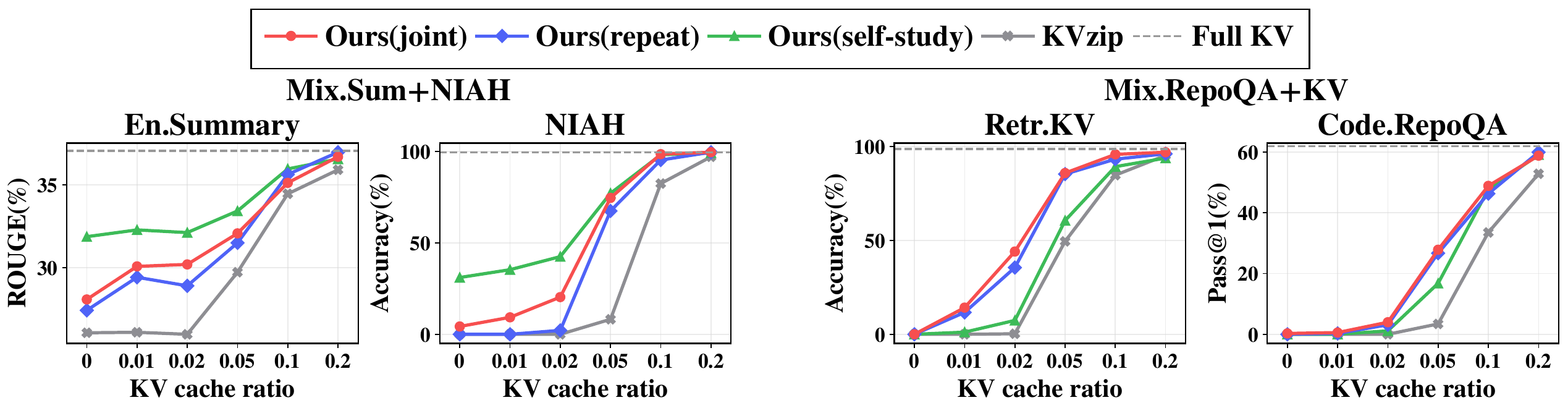}
    \caption{Multi-task results on RepoQA+KV and Summary+NIAH using Qwen2.5-7B-1M.}
    \label{fig:multi_task}
    \vspace{-1.5em}
\end{figure}

\textbf{Multi-task KV caches.}
To evaluate \ours{} beyond the single-task setting of Fig.~\ref{fig:main-kvzip}, we construct mixed contexts concatenated from multiple benchmarks and assess whether a single patch can serve different downstream tasks simultaneously.
In Fig.~\ref{fig:multi_task}, we report performance on two mixed contexts: RepoQA+KV, combining two retrieval tasks, and Summary+NIAH, combining long-document summarization with text retrieval.
For each mixture, we build a single compressed cache and a corresponding weight patch, then evaluate on the queries associated with each constituent task separately.
\ours{} consistently improves over the compressed baseline on both tasks, with the gain persisting across compression ratios.
This indicates that a single patch built jointly over a heterogeneous context is not dominated by one task at the expense of another.
We attribute this to two aspects of our design: the reference query set represents the query distribution regardless of task structure, and joint regression in Eq.~\eqref{eq:ridge} aggregates compression-induced errors from all regions into a single weight patch.
We also observe that reference query strategies transfer across task mixtures: self-study performs better on summary contexts, while repeat-context gains an edge when retrieval is present, consistent with our findings in Sec.~\ref{sec:main_results}.

\subsection{Ablation Study}

\label{sec:ablation}

\begin{wraptable}{r}{0.4\linewidth}
\centering
\small
\setlength{\tabcolsep}{3pt}
\caption{Effect of chunk size.}
\label{tab:chunk-size}
\begin{tabular}{lccccc}
\toprule
Chunk & 1K & 2K & \textbf{5K} & 8K & 10K \\
\midrule
Rel. Acc. & \textbf{.696} & .695 & \textbf{.696} & .672 & .679 \\
Time (s) & 114 & 98 & \textbf{85} & 86 & 81 \\
Mem. (GB) & 30.8 & 31.6 & \textbf{33.0} & 34.1 & 35.9 \\
\bottomrule
\end{tabular}
\vspace{-0.7em}
\end{wraptable}

\textbf{Effect of chunk size.}
We study how the chunk size $L_\text{chunk}$ used in patch construction (Sec.~\ref{sec:efficient_computation}) trades off patch quality against construction cost.
Fixing the cache budget at $b/n = 0.1$, we vary $L_\text{chunk}$ from 1K to 10K tokens and report average relative accuracy and patch construction overhead over the benchmark in Tab.~\ref{tab:chunk-size}.
Accuracy stays flat from 1K to 5K and slightly drops at 8K and beyond, consistent with the assumption in Sec.~\ref{sec:efficient_computation} that cross-chunk attention can be safely dropped only when relevant dependencies remain within a single chunk.
We attribute the mild accuracy drop at larger chunk sizes to the difficulty of long-span reference reconstruction, which yields more noised teacher activations and degrades the ridge target.
Time decreases sharply from 1K to 5K and saturates beyond that, while peak memory grows monotonically with $L_\text{chunk}$, as expected from the chunked-computation design.
We therefore use 5K as the default chunk size with the best balance.

\begin{wraptable}{r}{0.3\linewidth}
\vspace{-1.2em}
\centering
\small
\setlength{\tabcolsep}{3pt}
\caption{Ridge scaling results.}
\label{tab:ridge-scale}
\vspace{-0.1cm}
\begin{tabular}{ccc}
\toprule
Ratio & Trace &  Weighted \\
\midrule
0.2  & 0.92 & \textbf{0.93} \\
0.1  & 0.88 & \textbf{0.91}  \\
0.05 & 0.72 & 0.72  \\
0.02 & 0.45 & \textbf{0.46} \\
0.01 & 0.32 & \textbf{0.36} \\
\bottomrule
\end{tabular}
\vspace{-0.7em}
\end{wraptable}

\textbf{Effect of ridge scaling.} In Eq.~\eqref{eq:ridge}, $H^{\prime}$ and $T$ vary across layers and contexts, so the ridge term should be scaled relative to the layerwise regression matrix. We compare different layer-adaptive schemes that normalize $\lambda$ by spectral quantities of $S_H=H^{\prime\top}H^{\prime}$: trace mean $\mathrm{tr}(S_H)/d_{ff}$, and our variant $||S_H||^2_F/\mathrm{tr}(S_H)$, corresponding to the mean, and weighted mean of eigenvalues respectively.
Tab.~\ref{tab:ridge-scale} shows that our default $||S_H||^2_F/\mathrm{tr}(S_H)$ consistently performs better between 3 tasks of each group: NIAH, En.QA, and Math.Find, patched by self-study reference query. For each scaling strategy, we report the best performance obtained by selecting best $\lambda_0$. Our normalization provides an intermediate scale between the trace mean and spectral norm, emphasizing dominant directions without collapsing onto the largest eigenvalue. We empirically find that it transfers across contexts, models, and cache ratios without any retuning.

\begin{figure}[t]
    \centering
    \includegraphics[width=\linewidth]{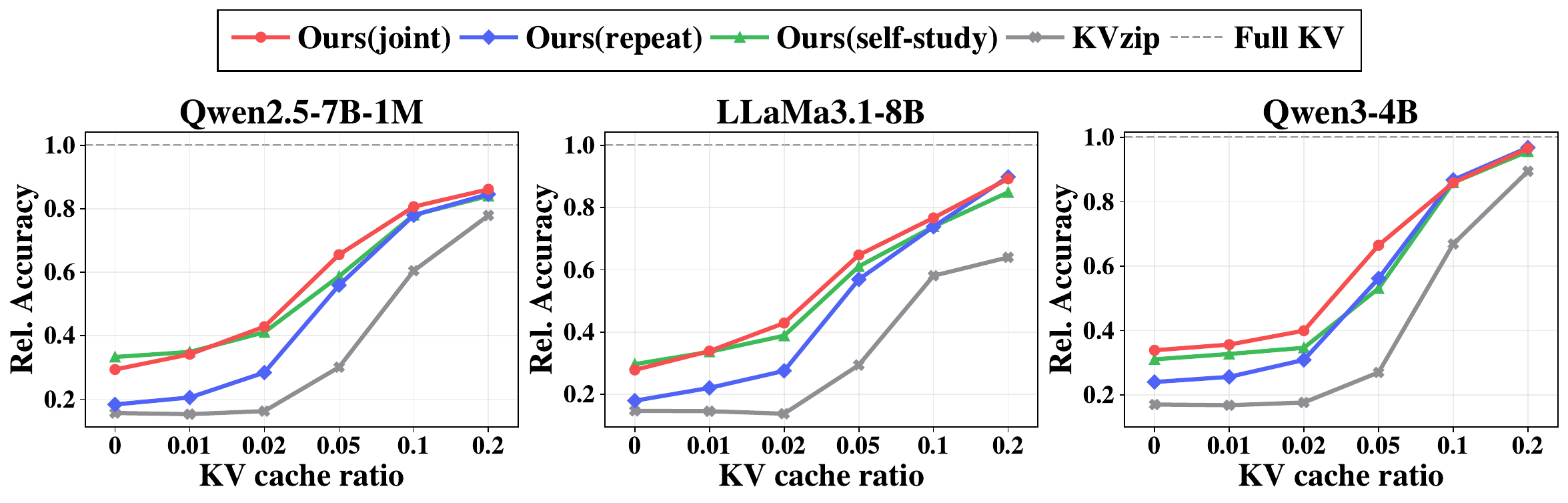}
    \caption{
    Averaged score across datasets, normalized by the full-KV score, for different models.
    }
    \label{fig:model_comparison}
    \vspace{-0.5cm}
\end{figure}

\textbf{Model scale and architectures.}
We also test \ours{} across model families and scales, including Qwen2.5-7B-1M~\cite{qwen2025qwen25technicalreport}, Qwen3-4B~\citep{yang2025qwen3}, and LLaMA3.1-8B~\cite{grattafiori2024llama}.
Fig.~\ref{fig:model_comparison} reports the average relative accuracy across datasets, where each score is normalized by the performance with full KV cache. 
Applied on top of KVzip, \ours{} consistently improves over the cache-only baseline across all models, with the largest gains in the aggressive cache regime ($b/n$ between 0.02 and 0.1) where the baseline degrades sharply. 
The relative ranking of reference query strategies is also stable across architectures: the joint and self-study variants are strongest at aggressive budgets, while repeat-context becomes competitive as more KV pairs are retained.
Together, these results suggest that \ours{} is largely architecture-agnostic, compensating for compression-induced information loss across models.
\section{Related Work}
\label{sec:related_work}

\textbf{KV cache compression.}
Token-eviction methods retain a subset of the cache by scoring tokens with attention statistics or reconstruction signals~\cite{li2024snapkv, zhang2023h2o, cai2025pyramidkv}, and are query-aware, requiring the downstream query at selection time.
KVzip~\cite{kim2025kvzip} extends this to the query-agnostic setting by scoring tokens through context reconstruction, producing a single compressed cache reusable across queries.
Attention Matching~\cite{zweiger2026fast} instead approximates the full cache by constructing a compact cache that reproduces full-cache attention outputs.
\ours{} pairs a compressed cache with a context-specific weight patch that compensates for compression-induced error, making it compatible with different compression methods.

\textbf{Compensating for compression error.}
Concurrent work has also explored compensating for information discarded during KV cache compression.
GRKV~\citep{peng2026grkv} redistributes information from evicted tokens into retained KV states, while MomentKV~\citep{li2026momentkv} maintains compact statistics of evicted states to approximate their residual attention contribution.
While these methods compensate for discarded information through KV cache or attention-side representations, \ours{} represents the residual correction through context-specific MLP weight updates.

\textbf{Converting context into weights.}
\citet{dherin2025learning} show that the effect of a context segment on a transformer block can be represented exactly by a low-rank MLP weight update, and subsequent works extend this to more general architectures~\cite{goldwaser2026equivalence} and sequential optimization over reference completions~\cite{mazzawi2025transmuting}.
Doc-to-LoRA~\cite{charakorn2026doctolora} learns a low-rank adapter that internalizes a document through context distillation at inference time.
These works largely treat the weight update as a stand-alone replacement for the context, often under simplified assumptions such as fixed query tokens or access to only a subset of the original context. \ours{} repurposes context-to-weight conversion for a more practical use case: rather than replacing the context, the patch complements an off-the-shelf compressed KV cache by closing its residual compression error in closed form, making the mechanism directly applicable to existing long-context serving pipelines.
\section{Conclusion}

We introduce \ours{}, a framework that complements KV cache compression with context-derived weight patches.
Given an off-the-shelf compressed cache, \ours{} constructs block-wise weight patches by aligning activations of reference queries with full cache and compressed cache, yielding a closed-form ridge-regression solution.
Experiments show that our method offers consistent improvements over compression methods at low-cache regime, across diverse benchmarks and models.
These results suggest that \ours{} provides a general complement to cache-only compression, enabling more aggressive KV cache compression without increasing per-query inference cost.

\begin{ack}
This work was in part supported by the National Research Foundation of Korea (RS-2024-00351212 and RS-2024-00436165), the Institute of Information \& communications Technology Planning \& Evaluation (IITP) (RS-2022-II220926, and RS-2022-II220959, RS-2026-25518317, RS-2019-II190075).
This work was also supported by the InnoCORE program of the MSIT (AI Meta-Scientist, N10260110), the “HPC support” project funded by the Korea government, and the Korea Meteorological Administration Research and Development Program "Developing AI Technology for Weather Forecasting" under Grant (KMA2021-00121).
\end{ack}

\bibliography{reference}
\bibliographystyle{unsrtnat}

\clearpage
\appendix
\newpage
\section{Algorithms}
\label{appendix:algorithm}
\begin{algorithm}[H]
\caption{\ours{} weight patch construction}
\label{alg:patchkv}
\label{alg:patch_construction}
\begin{algorithmic}[1]
\REQUIRE Context $c$, base model $f_\theta$ with $L$ blocks,
KV cache compressor, cache budget $b$, ridge coefficient $\lambda_0$,
chunk size $L_{\mathrm{chunk}}$

\STATE $\Delta\theta_c \leftarrow \emptyset$

\vspace{2pt}
\STATE \textbf{Cache preparation}
\STATE Prefill full cache $C \leftarrow f_\theta(c)$
\STATE Compress $C_b \leftarrow \mathrm{Compress}(C,b)$
\STATE Build reference set $Q_c$ (repeat-context / self-study / joint)
\STATE Partition $Q_c$ into chunks $\{q_j\}$
\STATE Initialize teacher and student states
      $x^{F}_{j,1}, x^{C}_{j,1}$ from each reference chunk $q_j$

\vspace{2pt}
\STATE \textbf{Block-wise patch construction}
\FOR{$\ell = 1$ \TO $L$}

    \STATE Initialize
    $S_H \leftarrow 0 \in \mathbb{R}^{d_{\mathrm{ff}}\times d_{\mathrm{ff}}}$,
    $S_T \leftarrow 0 \in \mathbb{R}^{d\times d_{\mathrm{ff}}}$

    \FOR{each reference chunk $q_j$}

        \STATE \textbf{Teacher forward:}
        forward block $\ell$ from $x^{F}_{j,\ell}$ with full cache $C$;
        collect $(h_i,z_i)$ and obtain $x^{F}_{j,\ell+1}$

        \STATE \textbf{Student forward:}
        forward block $\ell$ from $x^{C}_{j,\ell}$ with compressed cache $C_b$;
        collect $(h'_i,z'_i)$

        \STATE Compute targets
        \[
        t_i \leftarrow
        W_\ell(h_i-h'_i)+(z_i-z'_i)
        \]

        \STATE Stack
        $H'_j \leftarrow [h'_i]_i$,
        $T_j \leftarrow [t_i]_i$

        \STATE Accumulate
        \[
        S_H \mathrel{+}= H_j'^{\top}H'_j,
        \qquad
        S_T \mathrel{+}= T_j^{\top}H'_j
        \]

        \STATE Discard per-token activations $(h_i,z_i,h'_i,z'_i)$

    \ENDFOR

    \STATE \textbf{Patch computation:}
    \[
    \lambda \leftarrow
    \lambda_0
    \frac{\|S_H\|_F^2}{\operatorname{tr}(S_H)}
    \]

    \STATE
    \[
    \Delta W_\ell
    \leftarrow
    S_T(S_H+\lambda I)^{-1}
    \]

    \STATE Merge
    $W_\ell \leftarrow W_\ell+\Delta W_\ell$

    \STATE
    $\Delta\theta_c
    \leftarrow
    \Delta\theta_c\cup\{\Delta W_\ell\}$

    \FOR{each reference chunk $q_j$}
        \STATE \textbf{Student-state propagation:}
        forward the patched block $\ell$ from $x^{C}_{j,\ell}$ with $C_b$
        and store the output as $x^{C}_{j,\ell+1}$
    \ENDFOR

\ENDFOR

\RETURN Patched weights $\theta+\Delta\theta_c$ and compressed cache $C_b$
\end{algorithmic}
\end{algorithm}

\section{Further Details in Experimental Setup}
\label{appendix_setup}

\subsection{Implementation details.} 
\paragraph{KV cache compression.} As described in Sec.~\ref{sec:setup}, we initialized the compressed KV cache with existing methods, KVzip~\citep{kim2025kvzip} and Attention Matching~\citep{zweiger2026fast}. For KVzip, we follow their default setting with prefill chunk size 2000 and global head scoring. 

In Attention Matching, due to large memory and time cost in extremely long contexts such as SCBench~\citep{li2025scbench}, we select the key as \textsc{HighestAttentionKeys} that select KV cache key $C_k$ by scoring. For other hyperparameters, we follow the best reported setting of Attention Matching using \textsc{HighestAttentionKeys}.

\begin{figure}[t]
    \centering
    \includegraphics[width=0.38\linewidth]{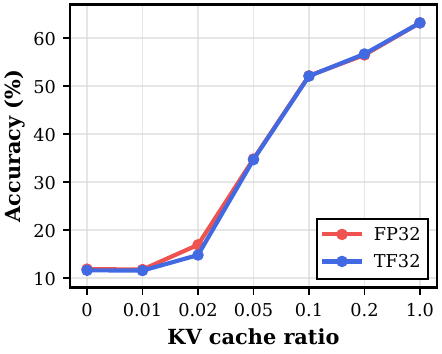}
    \caption{FP32 vs.\ TF32 patch construction.}
    \label{fig:numerical_precision}
\end{figure}

\begin{figure}[t]
    \centering
    \includegraphics[width=\linewidth]{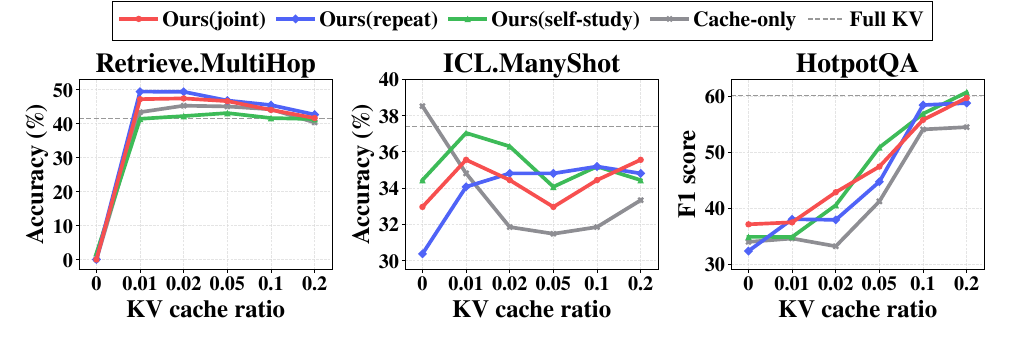}
    \caption{
    Additional benchmark results across KV cache budgets using Qwen2.5-7B-1M and KVzip.
    }
    \label{fig:additional_datasets}
\end{figure}

\textbf{Numerical Precision.} 
Patch construction involves accumulating sufficient statistics and solving the ridge regression in Eq.~\eqref{eq:ridge-solution}, which can be sensitive to numerical errors under low-precision computation. 
We therefore perform patch construction at higher precision than the BF16 precision used for the base model inference. 
Specifically, we keep patch construction in FP32 while enabling TF32 for the matrix multiplications involved in computing the sufficient statistics.
As shown in Fig.~\ref{fig:numerical_precision}, TF32 achieves downstream performance comparable to FP32 across KV cache budgets, while reducing patch construction time from 147.91\,s to 85.35\,s.
We therefore use TF32 matrix multiplication for patch construction throughout our experiments.

\begin{table}[t]
    \centering
    \small
    \caption{
    Score (\%) on the original SCBench retrieval tasks using Qwen2.5-7B-1M.
    Both methods achieve zero accuracy on \texttt{Retr.KV} and \texttt{Retr.Prefix-Suffix} under the evaluated cache budgets.
    }
    \label{tab:am_original_retrieval}
    \begin{tabular}{ll|ccccc}
        \toprule
        Task & Method & 0.20 & 0.10 & 0.05 & 0.02 & 0.01 \\
        \midrule
        \multirow{2}{*}{\texttt{Retr.KV}}
        & Attention Matching & 0.0 & 0.0 & 0.0 & 0.0 & 0.0 \\
        & + \ours{}          & 0.0 & 0.0 & 0.0 & 0.0 & 0.0 \\
        \midrule
        \multirow{2}{*}{\texttt{Retr.Prefix-Suffix}}
        & Attention Matching & 0.0 & 0.0 & 0.0 & 0.0 & 0.0 \\
        & + \ours{}          & 0.0 & 0.0 & 0.0 & 0.0 & 0.0 \\
        \midrule
        \multirow{2}{*}{\texttt{Code.RepoQA}}
        & Attention Matching & 15.23 & 3.18 & 1.59 & 0.23 & 0.23 \\
        & + \ours{}          & 33.64 & 8.64 & 3.41 & 0.91 & 0.68 \\
        \bottomrule
    \end{tabular}
\end{table}

\subsection{Dataset Selection and Task Variants}
\label{app:dataset_selection}

\paragraph{Dataset selection.}
SCBench includes nine tasks, of which seven are used in our main evaluation.
We report \texttt{Retrieve.MultiHop} and \texttt{ICL.ManyShot} separately since compression itself can improve performance on these tasks in these evaluation settings. This makes it difficult to distinguish the effect of \ours{} from the benefit of compression itself.
For \texttt{Retrieve.MultiHop}, which contains a large amount of distracting context, the evaluation results show that removing more context can improve performance by removing distractors. 
For \texttt{ICL.ManyShot}, we observe a more unusual pattern in which the baseline achieves its highest accuracy when no context KV pairs are retained, even exceeding the full-KV baseline.
This suggests that the many-shot context may not always be beneficial in this setting. One possible explanation is that the pretrained model can solve much of the task directly, while the provided context introduce additional distracting signals.
We report the results of \texttt{Retrieve.MultiHop} and \texttt{ICL.ManyShot} in Fig.~\ref{fig:additional_datasets}. 

Nevertheless, multi-hop reasoning remains an important setting for evaluating whether \ours{} remains effective when answering requires information from multiple parts of the context.
As shown in Fig.~\ref{fig:additional_datasets}, \ours{} with repeat-context or joint reference queries further improves over KVzip across several cache budgets in \texttt{Retrieve.MultiHop}.
To further verify that this result is not specific to the highly redundant synthetic structure of \texttt{Retrieve.MultiHop}, we additionally evaluate \ours{} on HotpotQA \citep{yang2018hotpotqa}, a natural multi-hop question answering benchmark.
\ours{} improves over KVzip across the evaluated nonzero cache budgets, providing complementary evidence that \ours{} remains effective on multi-hop reasoning tasks.

\paragraph{Task variants.}
For the Attention Matching experiments in Fig.~\ref{fig:main-kvzip}, we use shortened variants of the SCBench retrieval tasks.
On \texttt{Retr.KV} and \texttt{Retr.Prefix-Suffix}, Attention Matching collapses to zero accuracy under the evaluated budget, and applying \ours{} does not recover nonzero performance.
The original tasks therefore provide little resolution for evaluating the additional compensating effect of \ours{} over Attention Matching.
We instead evaluate the corresponding shortened variants, where the cache-only baseline retains nonzero performance under aggressive compression and the effect of weight-space compensation can be measured.
For completeness, we report the results on the original tasks in Tab.~\ref{tab:am_original_retrieval}.

\section{Additional Experiment Results.}

\label{appendix:add_exp}
\begin{table}[!t]
  \small
  \centering
  \caption{Stagewise wall-clock time and peak GPU memory for \ours{} construction.}
  \label{tab:runtime}
  \setlength{\tabcolsep}{8pt}
  \begin{tabular}{l|cc}
    \toprule
    \textbf{Stage} & \textbf{Time (s)} & \textbf{Peak Memory (GB)} \\
    \midrule
    (i) Cache compression (KVzip)      & 67.82 & 24.89\\
    (ii) Teacher forward & 41.41 & 30.92\\
    (iii) Student forward & 16.38 & 30.99 \\
    (iv) Patch computation & 11.40 & 32.98 \\
    (v) Student-state propagation & 16.16 & 31.11 \\
    \midrule
    Total                        & 153.17 & 32.98 \\
    \bottomrule
  \end{tabular}
\end{table}
\begin{figure}[t]
    \centering
    \includegraphics[width=0.5\linewidth]{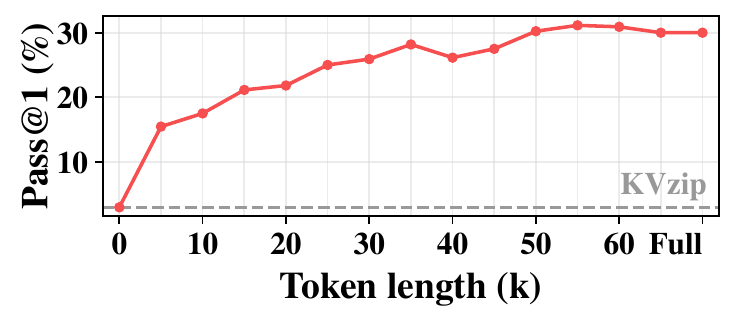}
    \caption{
    Performance across reference context lengths.
    }
    \label{fig:chunk_count}
\end{figure}
\textbf{Runtime Analysis.} We present a stagewise breakdown of the additional computational cost introduced by \ours{}. 
Patch construction consists of a one-time cache-compression stage, followed by four stages repeated for each transformer block: (ii) teacher forward, (iii) student forward, (iv) patch computation, and (v) student-state propagation. Wall-clock time and peak GPU memory for each stage are reported in Tab.~\ref{tab:runtime}, measured on a context of 169,035 tokens with repeat-context reference query and cache budget 0.2. For steps (ii)-(v), wall-clock time is aggregated for all transformer blocks.

\begin{figure}[t]
    \centering
    \includegraphics[width=0.5\linewidth]{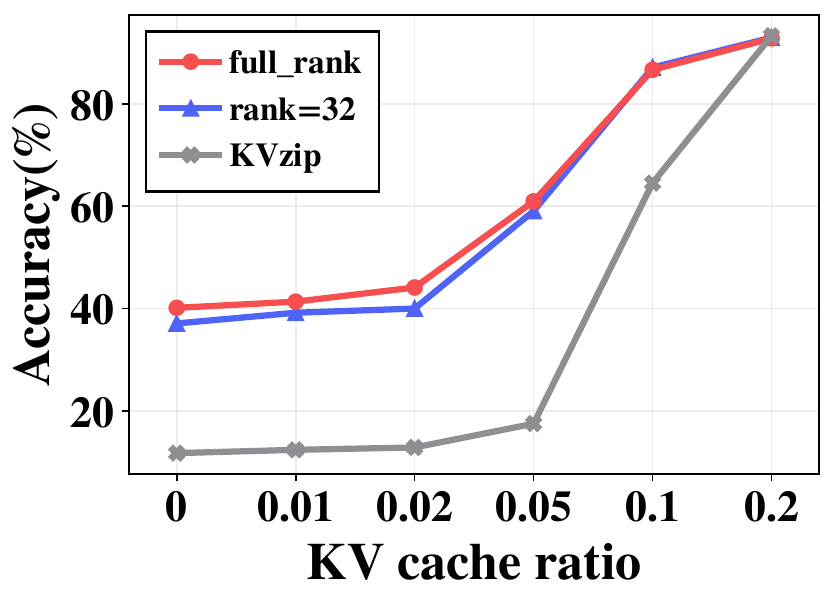}
    \caption{
    Performance when restricting the rank of weight patch of \ours{}.
    }
    \label{fig:lowrank}
\end{figure}

\textbf{Reference context budget.} In Fig.~\ref{fig:chunk_count}, we analyze the effect of limiting the amount of context used for patch construction on SCBench RepoQA, where the context length reaches up to 68K tokens. By controlling how many chunks of the context are fed as reference queries, we show that \ours{} achieves notable gains even with a small reference budget: using only 5K tokens (a single chunk) already recovers 15\% of the gap between the compressed and full-KV baseline. Performance continues to improve as more context is included, but with diminishing returns, suggesting that a modest reference budget is sufficient in practice and that \ours{}'s construction cost can be reduced significantly without sacrificing much accuracy.

\begin{figure}
    \centering
    \includegraphics[width=\linewidth]{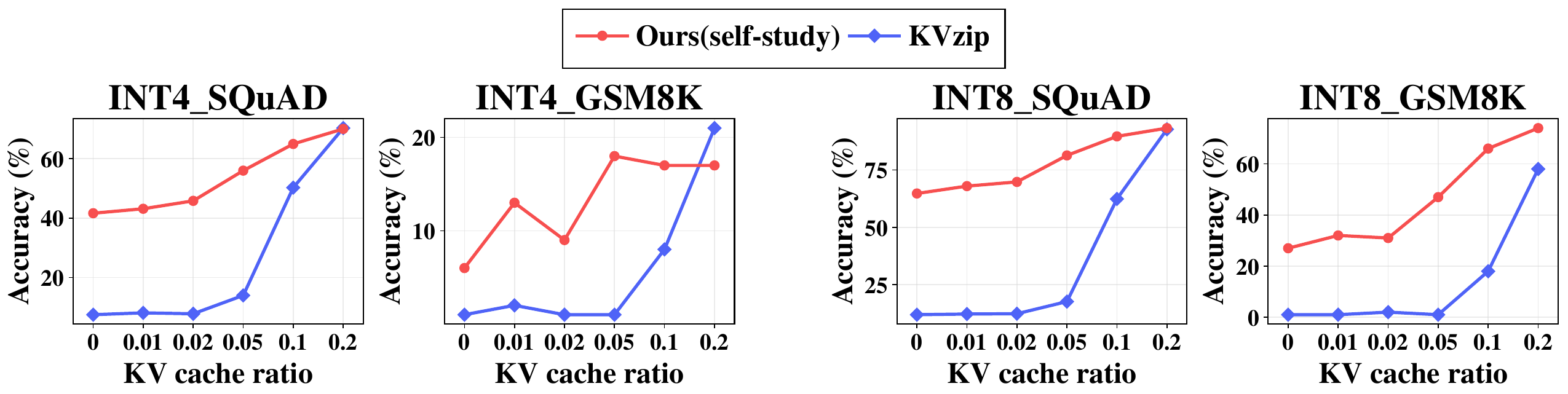}
    \caption{Performance of \ours{} on INT4/INT8 quantized KV cache.}
    \label{fig:quant}
\end{figure}

\paragraph{Compatibility with quantized KV caches.}
A complementary line of work reduces KV cache memory by quantizing it to lower precision~\citep{hooper2024kvquant, zandieh2026turboquant}. Since \ours{} operates as a residual update to the model weights and is agnostic to how the cache itself is stored, it can be combined with quantization without any modification. To verify this, we apply \ours{} on top of INT4 and INT8 quantized KV caches and report the results in Fig.~\ref{fig:quant}. \ours{} improves accuracy over the baseline at both INT8 and INT4 precision, with gains comparable to those observed in the unquantized setting. This confirms that \ours{} serves as an orthogonal solution to KV cache quantization for efficient LLM inference.

\paragraph{Low-rank weight patches.}
In some deployment scenarios, the weight patch may need to be stored alongside the base model, introducing additional memory overhead.
We therefore consider a low-rank variant of \ours{} that restricts the patch to rank $r$ (we set $r=32$).
Concretely, we apply a reduced-rank regression~\citep{izenman1975reduced} to the closed-form ridge solution, where we project $\Delta W_{\text{ridge}} = S_T (S_H + \lambda I)^{-1}$ onto its top-$r$ output subspace via $\Delta W_r = U_r U_r^\top \Delta W_{\text{ridge}}$. Here $U_r$ contains the top-$r$ eigenvectors of $\Delta W_{\text{ridge}} S_H \Delta W_{\text{ridge}}^\top$.
This keeps the closed-form structure intact and reduces per-context patch storage from $d \cdot d_{\text{ff}}$ to $(d + d_{\text{ff}}) \cdot r$.
In Fig.~\ref{fig:lowrank}, we compare the performance of full-rank \ours{} with the variant with weight patch with rank-$32$ on Qwen2.5-7B-1M, constructed by repeat-context reference query. As shown, the low-rank patch matches full-rank performance within a small margin while reducing per-context patch storage by $99\%$, demonstrating its practical viability.

\paragraph{Online Extension.}
\label{app:online_extension}

Our main experiments construct a context-specific patch once and reuse it across subsequent queries. We additionally examine whether weight-space compensation can be repeatedly applied as the KV cache evolves during generation.
Following the online compaction setting of Attention Matching (AM), we evaluate an online variant of \ours{} on AIME 2025 using Qwen3-4B, repeat-context reference queries.
During decoding, we compact the KV cache whenever it reaches $P=2048$ entries. At each event, we protect the most recent 20 entries and apply AM. \ours{} is then applied on top of the compacted cache before decoding resumes.

We evaluate generation limits of 4K and 8K tokens while varying the KV cache size after compaction. As shown in Tab.~\ref{tab:online_aime}, \ours{} provides larger gains under more aggressive compaction.
With an 8K and 4K generation limit, \ours{} show improvements compared to AM in more aggressive budgets, whereas \ours{} provides little or no improvement at larger post-compaction cache sizes.
These results provide initial evidence that weight-space compensation can be repeatedly applied as generated KV states are compressed during long reasoning. 
A broader evaluation of repeated updates and their long-horizon stability is left for future work.

\begin{table}[t]
    \centering
    \small
    \caption{
    Online evaluation on AIME 2025 using Qwen3-4B.
    A compaction event is triggered whenever the physical KV cache reaches
    2048 entries, with the most recent 20 entries protected during compaction.
    $2048 \rightarrow m$ denotes the post-compaction KV cache size.
    }
    \label{tab:online_aime}
    \begin{tabular}{c|c|cc}
        \toprule
        KV cache size & Method & 4K Acc. & 8K Acc. \\
        \midrule
        Full KV
        & Full
        & 9/30 & 14/30 \\
        \midrule

        \multirow{2}{*}{$2048 \rightarrow 1034$}
        & AM
        & 9/30 & \textbf{13/30} \\
        & AM + \ours{}
        & 9/30 & 12/30 \\
        \midrule

        \multirow{2}{*}{$2048 \rightarrow 527$}
        & AM
        & \textbf{9/30} & 11/30 \\
        & AM + \ours{}
        & 8/30 & \textbf{13/30} \\
        \midrule

        \multirow{2}{*}{$2048 \rightarrow 222$}
        & AM
        & 2/30 & 4/30 \\
        & AM + \ours{}
        & \textbf{8/30} & \textbf{8/30} \\
        \midrule

        \multirow{2}{*}{$2048 \rightarrow 121$}
        & AM
        & 2/30 & 2/30 \\
        & AM + \ours{}
        & \textbf{6/30} & \textbf{5/30} \\
        \bottomrule
    \end{tabular}
\end{table}

\paragraph{More qualitative examples.} From Fig.~\ref{fig:quali-2} to Fig.~\ref{fig:quali-5}, we show qualitative examples of query and answer with and without \ours{} on evicted context. 
\label{appendix:quan_exp}

\begin{figure}
    \centering
    \includegraphics[width=1.0\linewidth]{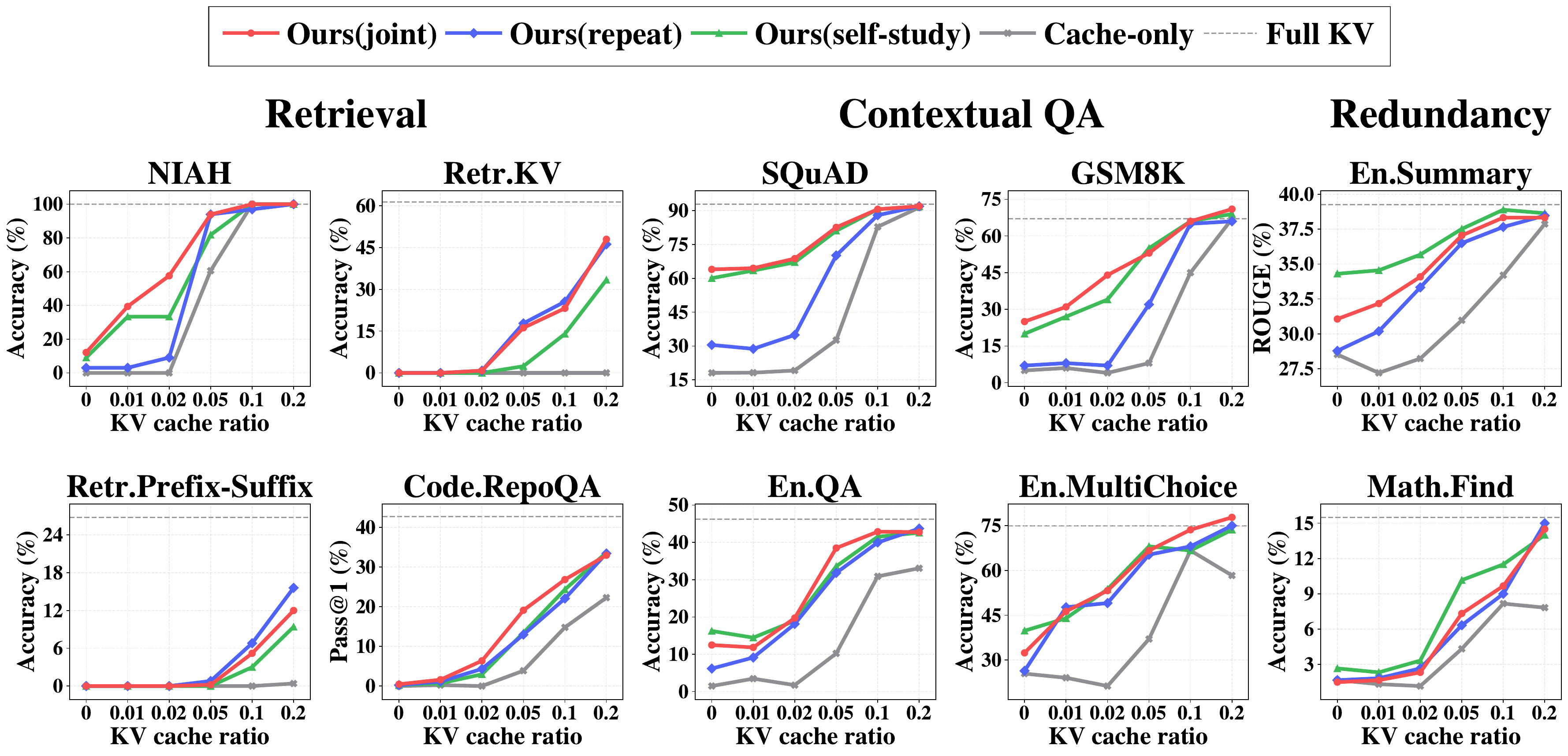}
    \caption{Benchmark results using LLaMA3.1-8B with \ours{} applied on top of an eviction based method (KVzip).}
    \label{fig:main-llama}
\end{figure}

\begin{figure}
    \centering
    \includegraphics[width=1.0\linewidth]{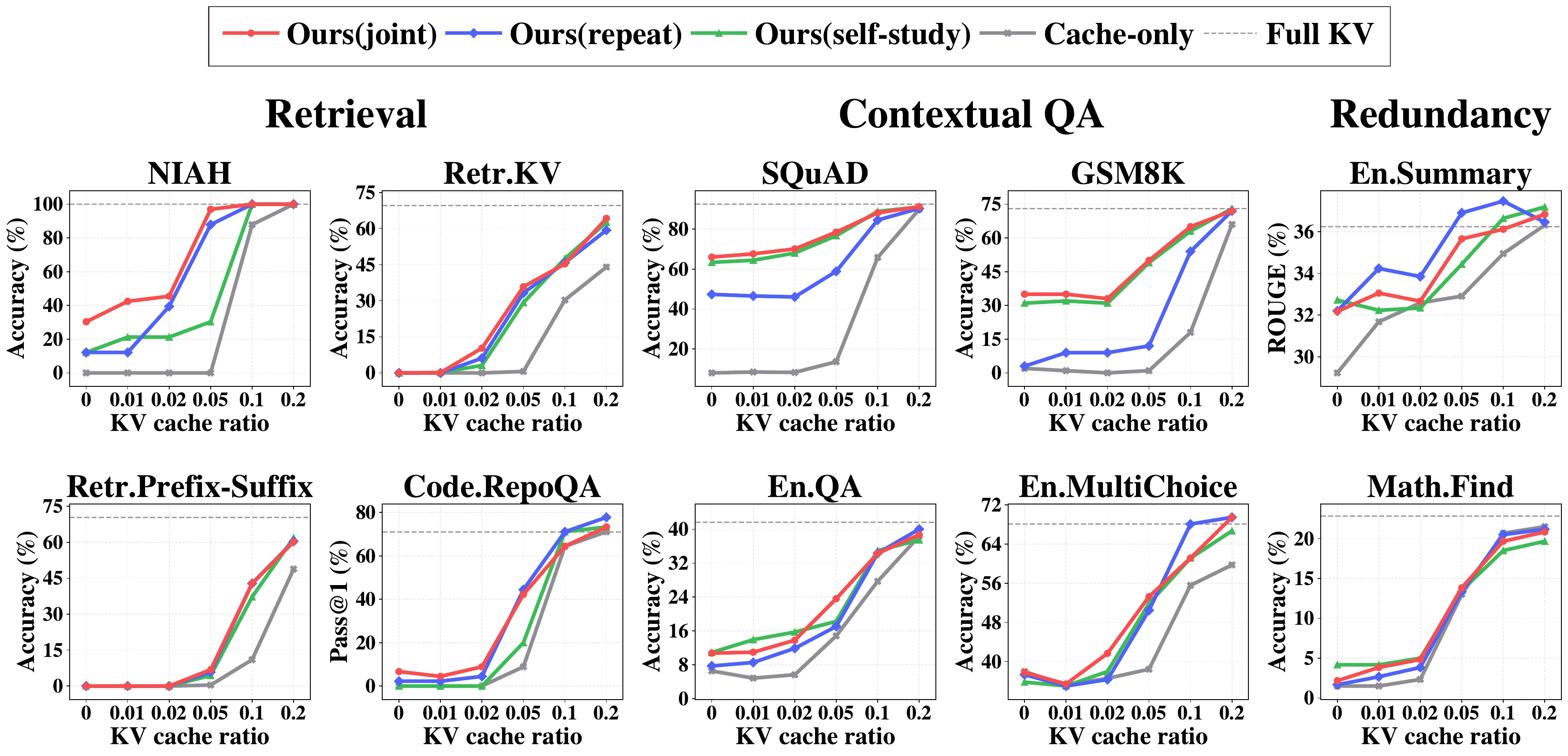}
    \caption{Benchmark results using Qwen3-4B with \ours{} applied on top of an eviction based method (KVzip).}
    \label{fig:main-qwen3}
\end{figure}

\paragraph{Full experiment results.}
In Fig.\ref{fig:main-llama} and Fig.\ref{fig:main-qwen3}, we report the full result of benchmarks evaluated in Qwen3-4B and LLaMA3.1-8B. Both models show similar property with Qwen2.5-7B-1M, consistently improving across all benchmarks and variants.

\section{Limitations}
\label{app:limitations}

\ours{} relies on context-derived reference queries as proxies for unknown future queries, so the reference-query mismatch may limit recovery under larger distribution shifts.
Patch construction also incurs a one-time preprocessing cost and access to the full KV cache as a teacher, making the method most suitable when this cost can be amortized across repeated queries or sufficiently long decoding.
Our primary evaluation targets the single-context, multi-query setting, while concurrent serving of many context-specific patches requires additional serving mechanisms.
The online results in Tab.~\ref{tab:online_aime} provide initial evidence beyond this static-context setting, but broader evaluation remains future work.

\begin{figure}
    \centering

\begin{prettybox}\scriptsize
    \textbf{Question:} Who currently provides uniforms to Notre Dame sport teams?
    \hfill \textcolor{gray}{\small \textsf{[SQuAD]}}

    \begin{contextbox}
    \textbf{Pruned Context:} \gdel{On July 1, 201}4,\gdel{ the} University of\gdel{ Notre Dame} and\kdel{Under Armour} reached an agreement in which \kkd{Under}{ Armour} will provide uniforms, apparel, equipment, and monetary compensation to Notre Dame for 1\gdel{0} years. [...] 
    \end{contextbox}

    {\color{black!20}\hrule height 0.8pt} 
    \vspace{1.5mm} 

    \noindent\textbf{Full KV:} Under Armour currently provides uniforms to Notre Dame sport teams. \hfill \textcolor{gray}{\small \textsc{Correct}}

    \vspace{1.5mm}
    {\color{black!10}\hrule height 0.5pt} 
    \vspace{1.5mm}

    \noindent\textbf{Pruned KV (w/o patch):} \textcolor{red}{Gymboree Uniforms} \hfill \textcolor{red}{\small \textsc{Incorrect}}

    \vspace{1.5mm}
    {\color{black!10}\hrule height 0.5pt} 
    \vspace{1.5mm}

    \noindent\textbf{PatchKV (Ours):} \textcolor{icebluetext}{Under Armour} currently provides uniforms to Notre Dame sport teams. \hfill \textcolor{icebluetext}{\small \textsc{Correct}}
\end{prettybox}

\caption{Qualitative example of \ours{} over KVzip on SQuAD. Evicted tokens shown in red.}
\label{fig:quali-2}
\end{figure}

\newpage
\vspace{5mm}
\begin{figure}
    \centering
    
\begin{prettybox}\scriptsize
    \textbf{Question:} Which other song from the soundtrack did better in the charts?
    \hfill \textcolor{gray}{\small \textsf{[SQuAD]}}

    \begin{contextbox}
    \textbf{Pruned Context:} [...] Bey\gdel{oncé released "F}ighting Tem\gdel{pt}ation" as\gdel{ the} lead\gdel{ single from} the film's soundtrack album, with Missy Elliott\gdel{,} MC Lyte, and Free which was also used to promote the film. Another of Beyoncé's contributions to the soundtrack, "\kkd{Sum}{mertime}", fared better on the US charts.
    \end{contextbox}

    {\color{black!20}\hrule height 0.8pt} \vspace{1.5mm} 

    \noindent\textbf{Full KV:} "Summertime” \hfill \textcolor{gray}{\small \textsc{Correct}}

    \vspace{1.5mm}{\color{black!10}\hrule height 0.5pt} \vspace{1.5mm}
    \vspace{1.5mm} 
    
    \noindent\textbf{Pruned KV (w/o patch):} I need more context to provide an accurate answer. \hfill \textcolor{red}{\small \textsc{Incorrect}}
    \\ Could you please specify which soundtrack you're referring to?

    \vspace{1.5mm}{\color{black!10}\hrule height 0.5pt} \vspace{1.5mm}

    \noindent\textbf{PatchKV (Ours):} \textcolor{icebluetext}{"Summertime”} \hfill \textcolor{icebluetext}{\small \textsc{Correct}}
    
\end{prettybox}
\caption{Qualitative example of \ours{} over KVzip on SQuAD. Evicted tokens shown in red.}
\label{fig:quali-3}
\end{figure}

\begin{figure}
    \centering
\begin{prettybox}\scriptsize
    \textbf{Question:} What is the most popsicle sticks he can make if he buys the cheapest lumber?
    \hfill \textcolor{gray}{\small \textsf{[GSM8K]}}

    \begin{contextbox}
    \textbf{Pruned Context:} Frederick is making popsicles to sell and to save money he is making his own popsicle sticks. He can get 200 sticks from a 2 x 4 piece of wood and 400 sticks from a {\setlength{\fboxsep}{1.5pt}\colorbox{hlred}{\textbf{\gdel{2} x \gdel{8}}}} piece of wood. He has \$24 to buy wood for sticks. A {\setlength{\fboxsep}{1.5pt}\colorbox{hlred}{\textbf{\gdel{2} x \gdel{4}}}} costs \$4. A \kdel{2 x 8} costs \$\kdel{6}.
    \end{contextbox}

    {\color{black!20}\hrule height 0.8pt} 
    \vspace{1.5mm} 

    \noindent\textbf{Full KV:} The answer is 1600 popsicle sticks. \hfill \textcolor{gray}{\small \textsc{Correct}}

    \vspace{1.5mm}
    {\color{black!10}\hrule height 0.5pt} 
    \vspace{1.5mm}
    
    \noindent\textbf{Pruned KV (w/o patch):} The most popsicle sticks he can make if he buys the cheapest lumber depends on the length \hfill \textcolor{red}{\small \textsc{Incorrect}}\\! and yield of the wood, but assuming 10 sticks per foot of 1x4 pine, a 10-foot piece would yield \textcolor{red}{100} popsicle sticks. 

    \vspace{1.5mm}
    {\color{black!10}\hrule height 0.5pt} 
    \vspace{1.5mm}

    \noindent\textbf{PatchKV (Ours):} The answer is that Frederick can make \textcolor{icebluetext}{1600} popsicle sticks if he buys the cheapest lumber. \hfill \textcolor{icebluetext}{\small \textsc{Correct}}
    
\end{prettybox}
\caption{Qualitative example of \ours{} over KVzip on GSM8K. Evicted tokens shown in red.}
\label{fig:quali-4} 
\end{figure}

\begin{figure}[h]
    \centering
\begin{prettybox}\scriptsize
    \textbf{Question:} What is Marcus' total bill?
    \hfill \textcolor{gray}{\small \textsf{[GSM8K]}}

    \begin{contextbox}
    \textbf{Pruned Context:} Marcus ordered 5 croissants at {\setlength{\fboxsep}{1.5pt}\colorbox{hlred}{\textbf{\$\gdel{3}.0\gdel{0}}}} apiece, 4 cinnamon rolls at \$2\gdel{.}50 each, 3 mini quiches for {\setlength{\fboxsep}{1.5pt}\colorbox{hlred}{\textbf{\$\gdel{4}.0\gdel{0}}}} apiece and 13 blueberry muffins that were \$1\gdel{.}00 apiece. At check out, Marcus shows his loyalty card that gives him {\setlength{\fboxsep}{1.5pt}\colorbox{hlred}{\textbf{\gdel{1}0\%}}} off of his purchase.
    \end{contextbox}

    {\color{black!20}\hrule height 0.8pt} 
    \vspace{1.5mm} 

    \noindent\textbf{Full KV:} The answer is Marcus' total bill is \$45.00. \hfill \textcolor{gray}{\small \textsc{Correct}}

    \vspace{1.5mm}
    {\color{black!10}\hrule height 0.5pt} 
    \vspace{1.5mm}
    
    \noindent\textbf{Pruned KV (w/o patch):} The answer is unknown based on the given information. \hfill \textcolor{red}{\small \textsc{Incorrect}}

    \vspace{1.5mm}
    {\color{black!10}\hrule height 0.5pt} 
    \vspace{1.5mm}

    \noindent\textbf{PatchKV (Ours):} The answer is \textcolor{icebluetext}{\$45.00}. \hfill \textcolor{icebluetext}{\small \textsc{Correct}}
    
\end{prettybox}
\caption{Qualitative example of \ours{} over KVzip on GSM8K. Evicted tokens shown in red.}
\label{fig:quali-5} 
\end{figure}

\end{document}